\documentclass[11pt]{article}
\usepackage{emnlp2016}
\usepackage{times}
\usepackage{latexsym}

\emnlpfinalcopy
\def\emnlppaperid{701}

\usepackage[utf8]{inputenc}
\usepackage{amssymb}
\usepackage{tikz}
\usepackage{multirow}
\usepackage{booktabs}
\usepackage{enumitem}
\usepackage{graphicx} 
\graphicspath{ {images/} }

\usepackage{amsmath}
\DeclareMathOperator{\softmax}{softmax}
\DeclareMathOperator{\float}{float}

\usepackage{xcolor}
\definecolor{nice-red}{HTML}{E41A1C}
\definecolor{nice-orange}{HTML}{FF7F00}
\definecolor{nice-yellow}{HTML}{FFC020}
\definecolor{nice-green}{HTML}{4DAF4A}
\definecolor{nice-blue}{HTML}{377EB8}
\definecolor{nice-purple}{HTML}{984EA3}

\usepackage[textsize=small]{todonotes}
\usepackage{comment}
\usepackage{enumitem}

\title{Numerically Grounded Language Models for Semantic Error Correction}
\author{
Georgios P. Spithourakis \and Isabelle Augenstein \and Sebastian Riedel \\
Deptartment of Computer Science \\ University College London \\
\{g.spithourakis, i.augenstein, s.riedel\}@cs.ucl.ac.uk
}

\begin{document}

\maketitle

\begin{abstract}

Semantic error detection and correction is an important task for applications such as fact checking, speech-to-text or grammatical error correction.
Current approaches generally focus on relatively shallow semantics and do not account for numeric quantities.
Our approach uses language models grounded in numbers within the text. Such groundings are easily achieved for recurrent neural language model architectures, which can be further conditioned on incomplete background knowledge bases.
Our evaluation on clinical reports shows that numerical grounding improves perplexity by 33\% and F1 for semantic error correction by 5 points when compared to ungrounded approaches. Conditioning on a knowledge base yields further improvements.

\end{abstract}

\section{Introduction}
\label{sec:intro}

\begin{figure}[!ht]
\centering
\includegraphics[width=0.48\textwidth]{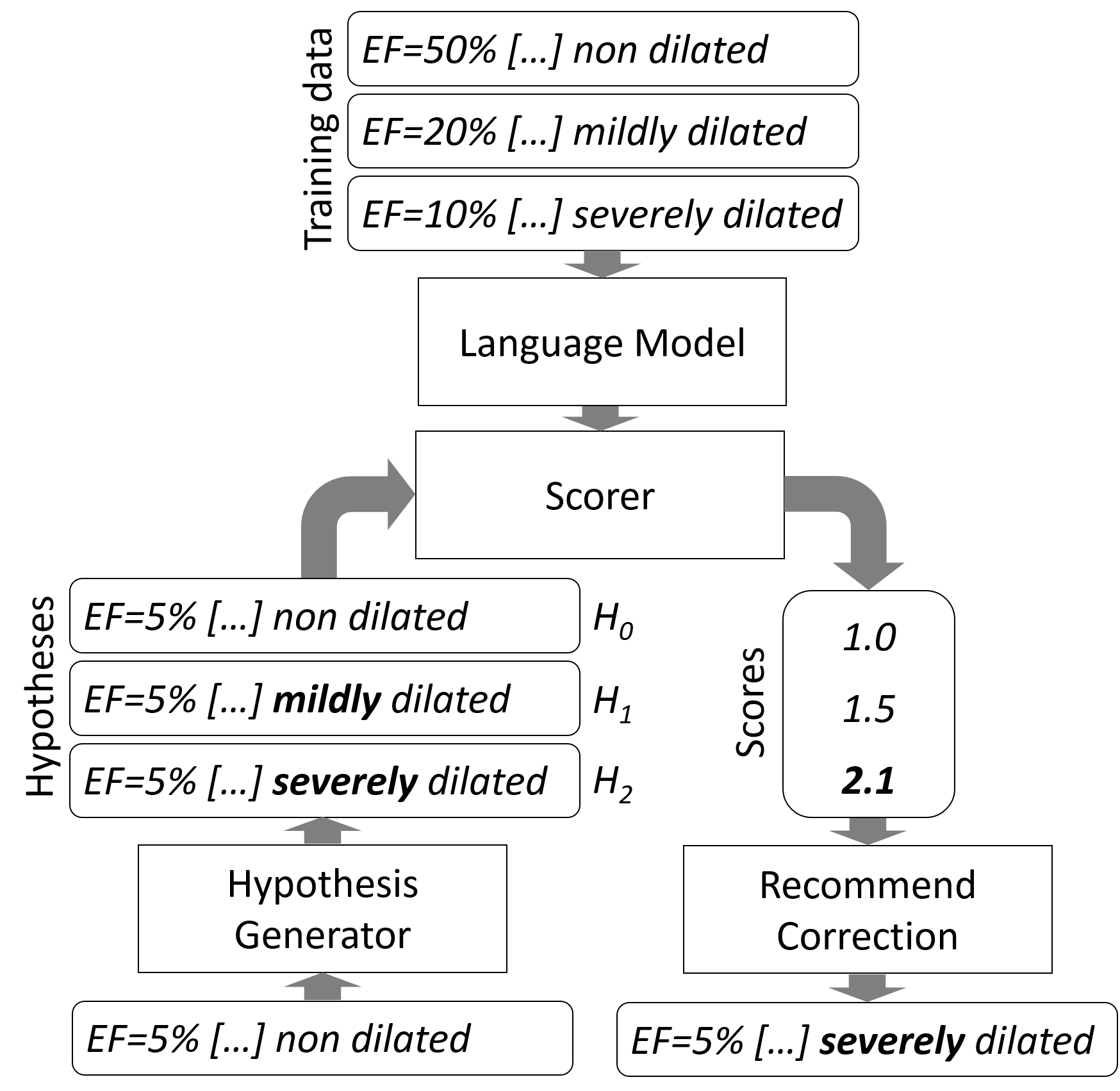}
\caption{Semantic error correction using language models.}
\label{fig:task_example}
\end{figure}

In many real world scenarios it is important to detect and potentially correct semantic errors and inconsistencies in text. For example, when clinicians compose reports, some statements in the text may be inconsistent with measurements taken from the patient~\cite{sue2013impact}.
Error rates in clinical data range from 2.3\% to 26.9\%~\cite{goldberg2008analysis} and many of them are number-based errors ~\cite{arts2002defining}.
Likewise, a blog writer may make statistical claims that contradict facts recorded in databases~\cite{munger2008blogging}. 
Numerical concepts constitute 29\% of contradictions in Wikipedia and GoogleNews~\cite{DeRaffertyEtAl2008} and 8.8\% of contradictory pairs in entailment datasets~\cite{dagan2006pascal}.

These inconsistencies may stem from oversight, lack of reporting guidelines or negligence. In fact they may not even be errors at all, but point to interesting outliers, or to errors in a reference database. In all cases, it is important to spot and possibly correct such inconsistencies. This task is known as semantic error correction (\textit{SEC})~\cite{DahlmeierNg2011}. 

In this paper, we propose a SEC approach to support clinicians with writing patient reports. A SEC system reads a patient's structured background information from a knowledge base (\textit{KB}) and their clinical report. Then it recommends improvements to the text of the report for semantic consistency. An example of an inconsistency is shown in Figure~\ref{fig:task_example}.
The SEC system has been trained on a dataset of records and learnt that the phrases ``non dilated'' and ``severely dilated'' correspond to high and low values for ``EF'' (abbreviation for ``ejection fraction'', a clinical measurement), respectively. If then the system is presented with the phrase ``non dilated'' in the context of a low value, it will detect a semantic inconsistency and correct the text to ``severely dilated''.

Our contributions are:
1) a straightforward extension to recurrent neural network (RNN) language models for \textit{grounding} them in numbers available in the text;
2) a simple method for modelling text \textit{conditioned on} an incomplete KB by lexicalising it; 
3) our evaluation on a semantic error correction task for clinical records shows that our method achieves F1 improvements of 5 and 6 percentage points with grounding and KB conditioning, respectively, over an ungrounded approach (F1 of 49\%).



\section{Methodology}
\label{sec:models}


Our approach to semantic error correction (Figure~\ref{fig:task_example}) starts with training a language model (LM), which can be grounded in numeric quantities mentioned in-line with text (Subsection~\ref{sec:Grounding}) and/or conditioned on a potentially incomplete KB (Subsection~\ref{sec:Conditioning}). Given a document for semantic checking, a hypothesis generator proposes corrections, which are then scored using the trained language model (Subsection~\ref{sec:SEC_with_LMs}). A final decision step involves accepting the best scoring hypothesis.


\begin{figure}[!ht]
\centering
\includegraphics[width=0.48\textwidth]{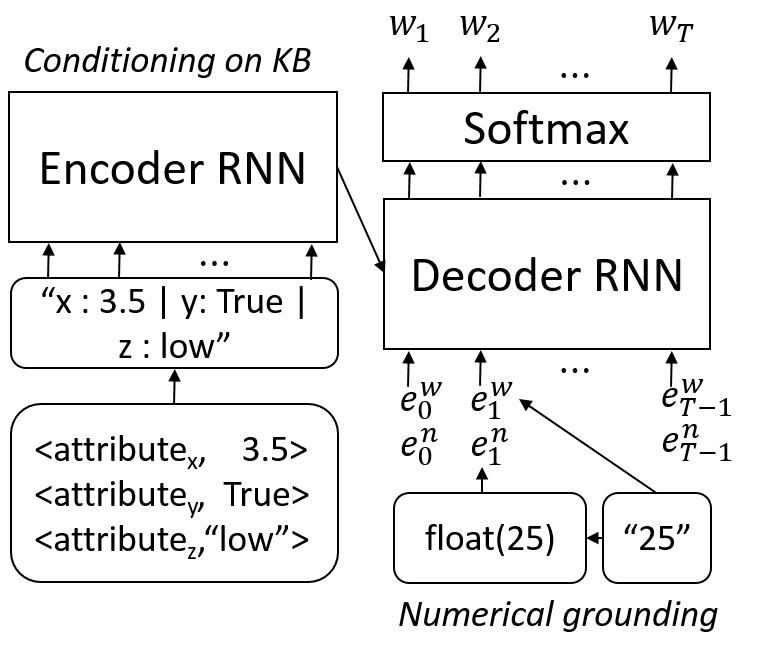}
\caption{A language model that is numerically grounded and conditioned on a lexicalised KB. Examples of data in rounded rectangles.}
\label{fig:system}
\end{figure}

\subsection{Numerically grounded language modelling}\label{sec:Grounding}


Let $\{w_1, ..., w_T\}$ denote a document, where $w_t$ is the one-hot representation of the $t$-th token and $V$ is the vocabulary size. 
A neural LM uses a matrix, $E_{in} \in \mathbb{R}^{D \times V}$, to derive word embeddings, $e^w_t = E_{in}w_t$.
A hidden state from the previous time step, $h_{t-1}$, and the current word embedding, $e^w_t$, are sequentially fed to an RNN's recurrence function to produce the current hidden state, $h_t \in \mathbb{R}^D$. The conditional probability of the next word is estimated as $\softmax(E_{out}h_t)$, where $E_{out} \in \mathbb{R}^{V \times D}$ is an output embeddings matrix.

We propose concatenating a representation, $e^n_t$, of the numeric value of $w_t$ to the inputs of the RNN's recurrence function at each time step.
Through this numeric representation, the model can generalise to out-of-vocabulary numbers.
A straightforward representation is defining $e^n_t=\float(w_t)$, where $\float(.)$ is a numeric conversion function that returns a floating point number constructed from the string of its input. If conversion fails, it returns zero.
The proposed mechanism for \emph{numerical grounding} is shown in Figure~\ref{fig:system}.
Now the probability of each next word depends on numbers that have appeared earlier in the text. 
We treat numbers as a separate modality that happens to share the same medium as natural language (text), but can convey exact measurements of properties of the real world. At training time, the numeric representations mediate to ground the language model in the real world.


\subsection{Conditioning on incomplete
KBs}\label{sec:Conditioning}

The proposed extension can also be used in \emph{conditional language modelling} of documents given a knowledge base. 
Consider a set of KB tuples accompanying each document and describing its attributes in the form $<attribute, value>$, where attributes are defined by a KB schema.
We can \emph{lexicalise} the KB by converting its tuples into textual statements of the form "$attribute : value$".
An example of how we lexicalise the KB is shown in Figure~\ref{fig:system}.
The generated tokens can then be interpreted for their word embeddings and numeric representations.
This approach can incorporate KB tuples flexibly, even when values of some attributes are missing.






\subsection{Semantic error correction}\label{sec:SEC_with_LMs}
A statistical model chooses the most likely correction from a set of possible correction choices.
If the model scores a corrected hypothesis higher than the original document, the correction is accepted.


A \emph{hypothesis generator function}, $G$, takes the original document, $H_0$, as input and generates a set of candidate corrected documents $G(H_0) = \{ H_1,...,H_M \}$. 
A simple hypothesis generator uses confusion sets of semantically related words to produce all possible substitutions.


A \emph{scorer model}, $s$, assigns a score $s(H_i) \in \mathbb{R}$ to a hypothesis $H_i$. The scorer is based on a likelihood ratio test between the original document (null hypothesis, $H_0$) and
each candidate correction (alternative hypotheses, $H_i$),
i.e. $s(H_i)=\frac{p(H_i)}{p(H_0)}$.
The assigned score represents how much more probable a correction is than the original document. 


The probability of observing a document, $p(H_i)$, can be estimated using language models, or grounded and conditional variants thereof.

\section{Data}
\label{sec:data}
\begin{table}
\centering
 \begin{tabular}{r r r r r r} 

 \hline
 & & & \bf train & \bf dev & \bf test \\ [0.5ex]
 \hline
 \multicolumn{3}{r}{\#documents} & 11,158 & 1,625 & 3,220 \\
 \hline
 
 \parbox[t]{2mm}{\multirow{3}{*}{\rotatebox[origin=c]{90}{\#tokens/}}}
 &
 \parbox[t]{2mm}{\multirow{3}{*}{\rotatebox[origin=c]{90}{doc}}}
 & all & 204.9 & 204.4 & 202.2\\
 & & words & 95.7\% & 95.7\% & 95.7\%\\
 & & numeric & 4.3\% & 4.3\% & 4.3\%\\
 \hline
 
 \parbox[t]{2mm}{\multirow{3}{*}{\rotatebox[origin=c]{90}{\#unique}}}
 &
 \parbox[t]{2mm}{\multirow{3}{*}{\rotatebox[origin=c]{90}{tokens}}}
 & all & 18,916 & 6,572 & 9,515\\
 & & words & 47.8\% & 58.25\% & 54.1\%\\
 & & numeric & 52.24\% & 41.9\% & 45.81\%\\
 \hline
 \parbox[t]{2mm}{\multirow{3}{*}{\rotatebox[origin=c]{90}{OOV}}}
 &
 \parbox[t]{2mm}{\multirow{3}{*}{\rotatebox[origin=c]{90}{rate}}}
 & all & 5.0\% & 5.1\% & 5.2\%\\
 & & words & 3.4\% & 3.5\% & 3.5\%\\
 & & numeric & 40.4\% & 40.8\% & 41.8\%\\
 \hline

 \end{tabular}
 \caption{Statistics for clinical dataset. Counts for non-numeric (\textit{words}) and \textit{numeric} tokens reported as percentage of counts for \textit{all} tokens. Out-of-vocabulary (OOV) rates are for vocabulary of $1000$ most frequent words in the train data.
 }
 \label{tab:data}
\end{table}


Our dataset comprises 16,003 clinical records from the London Chest Hospital (Table~\ref{tab:data}). Each patient record consists of a text report and accompanying structured KB tuples.
The latter describe 20 possible numeric attributes (age, gender, etc.), which are also partly contained in the report.
On average, $7.7$ tuples are completed per record.
Numeric tokens constitute only a small proportion of each sentence (4.3\%), but account for a large part of the unique tokens vocabulary ($>$40\%) and suffer high OOV rates.


\begin{table}
\centering
 \begin{tabular}{r l} 

 \hline
 \bf description & \bf confusion set \\ [0.5ex]
 \hline
 intensifiers (adv): & \textit{non}, \textit{mildly}, \textit{severely} \\ 
 intensifiers (adj):  & \textit{mild}, \textit{moderate}, \textit{severe}  \\
 units: & \textit{cm}, \textit{mm}, \textit{ml}, \textit{kg}, \textit{bpm} \\
 viability: & \textit{viable}, \textit{non-viable} \\
 quartiles: & \textit{25}, \textit{50}, \textit{75}, \textit{100} \\
 inequalities: & \textit{$<$}, \textit{$>$} \\
 \hline
\end{tabular}
\caption{Confusion sets.}
\label{tab:confusion_sets}
\end{table}

To evaluate SEC, we generate a ``corrupted'' dataset of semantic errors from the test part of the ``trusted'' dataset (Table~\ref{tab:data}, last column).
We manually build confusion sets (Table~\ref{tab:confusion_sets}) by searching the development set for words related to numeric quantities and grouping them if they appear in similar contexts.
Then, for each document in the trusted test set we generate an erroneous document by sampling a substitution from the confusion sets.
Documents with no possible substitution are excluded.
The resulting ``corrupted'' dataset is balanced, containing 2,926 correct and 2,926 incorrect documents.

\section{Results and discussion}
\label{sec:experiments}

Our \emph{base LM} is a single-layer long short-term memory network (LSTM,~\newcite{hochreiter1997long} with all latent dimensions (internal matrices, input and output embeddings) set to $D=50$.
We extend this baseline to a \emph{conditional} variant by conditioning on the lexicalised KB (see Section~\ref{sec:Conditioning}).
We also derive a numerically \emph{grounded} model by concatenating the numerical representation of each token to the inputs of the base LM model (see Section~\ref{sec:Grounding}). Finally, we consider a model that is both grounded and conditional (\emph{g-conditional}).

The vocabulary contains the $V=1000$ most frequent tokens in the training set. Out-of-vocabulary tokens are substituted with $<$\textit{num\_unk}$>$, if numeric, and $<$\textit{unk}$>$, otherwise. We extract the numerical representations before masking, so that the grounded models can generalise to out-of-vocabulary numbers.
Models are trained to minimise token cross-entropy,
with $20$ epochs of back-propagation and adaptive mini-batch gradient descent (AdaDelta)~\cite{journals/corr/abs-1212-5701}. 

For SEC, we use an oracle hypothesis generator that has access to the groundtruth confusion sets (Table~\ref{tab:confusion_sets}).
We estimate the scorer (Section~\ref{sec:SEC_with_LMs}) using the trained \emph{base}, \emph{conditional}, \emph{grounded} or \emph{g-conditional} LMs.
As additional baselines we consider a scorer that assigns \emph{random} scores from a uniform distribution and \emph{always} (\emph{never}) scorers that assign the lowest (highest) score to the original document and uniformly random scores to the corrections.


\begin{table}
\centering
 \begin{tabular}{r r r r} 

 \hline
 \bf model & \bf tokens & \bf PP & \bf APP \\ [0.5ex]
 \hline
 \multirow{3}{*}{base LM}
 & all & 14.96 & 22.11 \\
 & words & 13.93 & 17.94 \\
 & numeric & 72.38 & 2289.47 \\
  \hline
 \multirow{3}{*}{conditional}
 & all & 14.52 & 21.47  \\
 & words & 13.49  & 17.38\\
 & numeric & 74.48  & 2355.77 \\
 \hline
 \multirow{3}{*}{grounded}
 & all & 9.91 & 14.66 \\
 & words & 9.28 & 11.96 \\
 & numeric & 42.67 & 1349.59 \\
 \hline
 \multirow{3}{*}{g-conditional} 
 & all & \textbf{9.39} & \textbf{13.88} \\
 & words & \textbf{8.80}  & \textbf{11.33}  \\
 & numeric & \textbf{39.84}  & \textbf{1260.28} \\
 \hline
 
\end{tabular}
\caption{Language modelling evaluation results on the test set. We report perplexity (PP) and adjusted perplexity (APP).
Best results in \textbf{bold}.
}
\label{tab:entropy_results}
\end{table}


\subsection{Experiment 1: Numerically grounded LM}\label{sec:Results:sec:Grounding}

We report perplexity and adjusted perplexity~\cite{ueberla1994analysing} of our LMs on the test set for all tokens and token classes (Table~\ref{tab:entropy_results}).
Adjusted perplexity is not sensitive to OOV-rates and thus allows for meaningful comparisons across token classes.
Perplexities are high for numeric tokens because they form a large proportion of the vocabulary.
The \emph{grounded} and \emph{g-conditional} models achieved a 33.3\% and 36.9\% improvement in perplexity, respectively, over the \emph{base LM} model. Conditioning without grounding yields only slight improvements, because most of the numerical values from the lexicalised KB are out-of-vocabulary.

\begin{figure}
\centering

\includegraphics[width=0.5\textwidth]{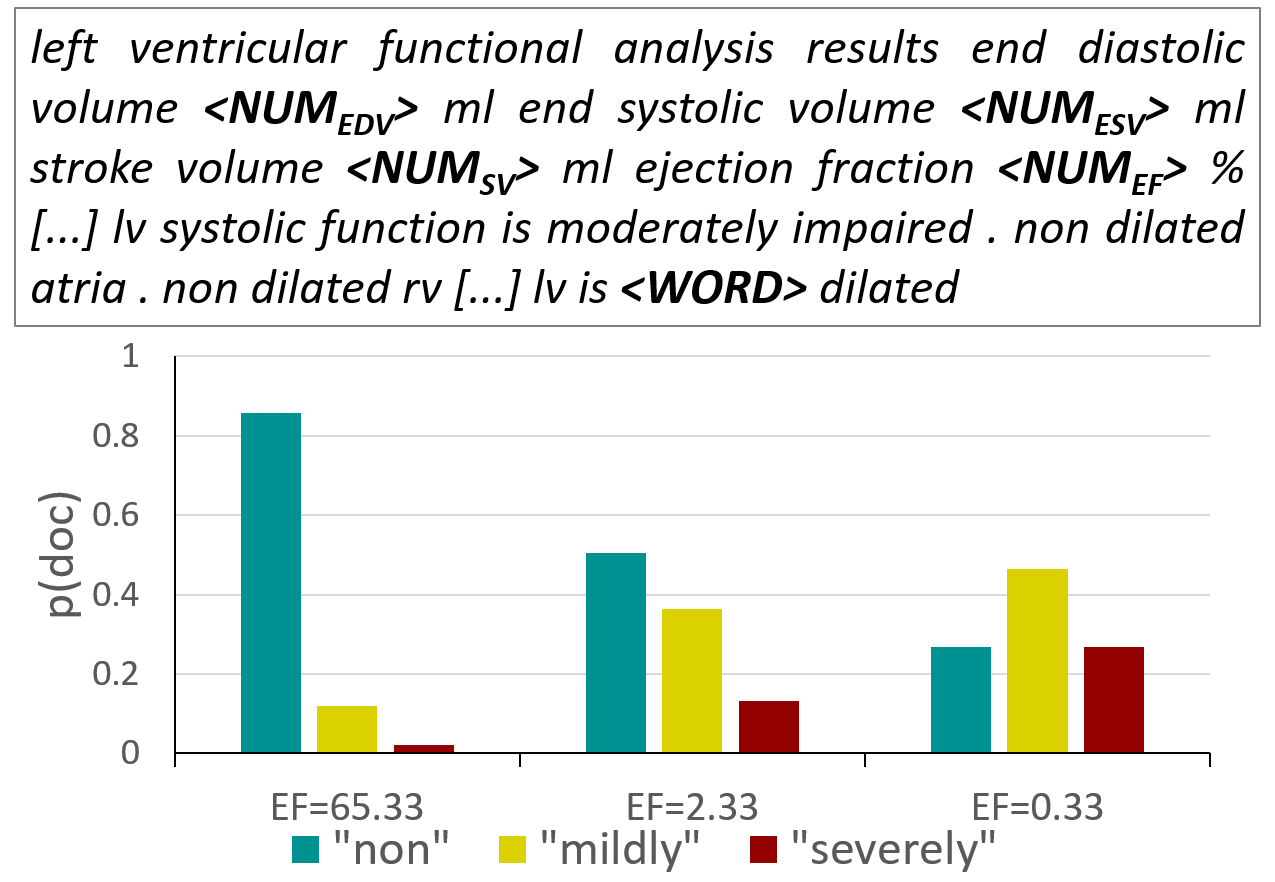}
\caption{Qualitative example. Template document and document probabilities for $<$WORD$>$=\{`non', `mildly', `severely'\} and varying numbers. Probabilities are renormalised over the set of possible choices.}
\label{fig:conditional_example}
\end{figure}

The qualitative example in 
Figure~\ref{fig:conditional_example} demonstrates how numeric values influence the probability of tokens given their history.
We select a document from the development set and substitute its numeric values as we vary $EF$ (the rest are set by solving a known system of equations). The selected exact values were unseen in the training data.
We calculate the probabilities for observing the document with different word choices \{``non'',  ``mildly'',  ``severely''\} under the \emph{grounded} LM and find that ``non dilated'' is associated with higher $EF$ values. This shows that it has captured semantic dependencies on numbers.


\subsection{Experiment 2: Semantic error correction} 

We evaluate SEC systems on the corrupted dataset (Section~\ref{sec:data}) for detection and correction.

For detection, we report precision, recall and F1 scores in Table~\ref{tab:detection_results}.
Our \emph{g-conditional} model achieves the best results, a total F1 improvement of 2 points over the \emph{base LM} model and 7 points over the best baseline. The conditional model without grounding performs slightly worse in the F1 metric than the \emph{base LM}.
Note that with more hypotheses the \emph{random} baseline behaves more similarly to \emph{always}. Our hypothesis generator generated on average 12 hypotheses per document. The results of \emph{never} are zero as it fails to detect any error.



For correction, we report mean average precision (MAP) in addition to the same metrics as for detection (Table~\ref{tab:correction_results}). The former measures the position of the ranking of the correct hypothesis. The \emph{always} (\emph{never}) baseline ranks the correct hypothesis at the top (bottom). Again, the \emph{g-conditional} model yields the best results, achieving an improvement of 6 points in F1 and 5 points in MAP over the \emph{base LM} model and an improvement of 47 points in F1 and 9 points in MAP over the best baseline. The \emph{conditional} model without grounding has the worst performance among the LM-based models.

\begin{table}[t!]
\centering
 \begin{tabular}{r r r r} 

 \hline
 \bf model & \bf P & \bf R & \bf F1 \\ [0.5ex]
 \hline
 random & 50.27 & 90.29 & 64.58 \\ 
 always & 50.00 & 100.0 & 66.67 \\ 
 never & 0.0 & 0.0 & 0.0 \\ 
 \hline
 base LM & 57.51 & 94.05 & 71.38 \\ 
 conditional & 56.86 & 94.43 & 70.98 \\
 grounded & 58.87 & 94.70 & 72.61 \\
 g-conditional & \textbf{60.48} & \textbf{95.25} & \textbf{73.98}\\ 
 \hline
\end{tabular}
\caption{Error detection results on the test set. We report precision (P), recall (R) and F1. Best results in \textbf{bold}.} 
\label{tab:detection_results}
\end{table}



\section{Related Work}
\label{sec:relatedwork}

Grounded language models represent the relationship between words and the non-linguistic context they refer to.
Previous work grounds language on 
vision~\cite{bruni2014multimodal,journals/tacl/SocherKLMN14,silberer-lapata:2014:P14-1},
audio~\cite{kiela-clark:2015:EMNLPa},
video~\cite{fleischman2008grounded},
colour~\cite{mcmahan2015bayesian},
and olfactory perception~\cite{kiela-bulat-clark:2015:ACL-IJCNLP}.
However, no previous approach has explored in-line numbers as a source of grounding.

Our language modelling approach to SEC is inspired by LM approaches to grammatical error detection (GEC)~\cite{NgWuEtAl2013,FeliceYuanEtAl2014}.
They similarly derive confusion sets of semantically related words, substitute the target words with alternatives and score them with an LM.
Existing semantic error correction approaches aim at correcting word error choices~\cite{DahlmeierNg2011}, collocation errors~\cite{kochmar2016error}, and semantic anomalies in adjective-noun combinations~\cite{vecchi2011linear}. So far, SEC approaches focus on short distance semantic agreement, whereas our approach can detect errors which require to resolve long-range dependencies. Work on GEC and SEC shows that language models are useful for error correction, however they neither ground in numeric quantities nor incorporate background KBs.

\begin{table}[t!]
\centering
 \begin{tabular}{r r r r r} 

 \hline
  \bf model & \bf MAP & \bf P & \bf R & \bf F1 \\ [0.5ex]
 \hline
 random  & 27.75 & 5.73 & 10.29 & 7.36 \\ 
 always  & 20.39 & 6.13 & 12.26 & 8.18 \\ 
 never  & 60.06 & 0.0  & 0.0 & 0.0 \\ 
 \hline
 base LM & 64.37 & 39.54 & 64.66 & 49.07 \\ 
 conditional & 62.76 & 37.46 & 62.20 & 46.76 \\
 grounded & 68.21 & 44.25 & 71.19 & 54.58 \\
 g-conditional & \textbf{69.14} & \textbf{45.36} & \textbf{71.43} & \textbf{55.48} \\ 
 \hline
\end{tabular}
\caption{Error correction results on the test set. We report mean average precision (MAP), precision (P), recall (R) and F1. Best results in \textbf{bold}.}
\label{tab:correction_results}
\end{table}


\section{Conclusion}
\label{sec:conclusion}

In this paper, we proposed a simple technique to model language in relation to numbers it refers to,
as well as conditionally on incomplete knowledge bases. We found that the proposed techniques lead to performance improvements in the tasks of language modelling, and semantic error detection and correction. Numerically grounded models make it possible to capture semantic dependencies of content words on numbers.

In future work, we will plan to apply numerically grounded models to other tasks, such as numeric error correction. We will explore alternative ways for deriving the numeric representations, such as accounting for verbal descriptions of numbers. For SEC, a trainable hypothesis generator can potentially improve the coverage of the system.


\ifemnlpfinal
\section*{Acknowledgments}
The authors would like to thank the anonymous reviewers for their insightful comments.
We also thank Steffen Petersen for providing the dataset and advising us on the clinical aspects of this work. This research was supported by the Farr Institute of Health Informatics Research.
\fi

\bibliographystyle{emnlp2016}
\bibliography{bibref}

\end{document}